\documentclass{llncs}
\usepackage[utf8]{inputenc}
\usepackage{amsmath}
\usepackage{amssymb}
\usepackage{mathtools}
\usepackage{multirow}
\usepackage{stmaryrd}
\usepackage{multirow}
\usepackage{etoolbox}
\usepackage[hyphens]{url}
\usepackage[breaklinks]{hyperref}
\usepackage{graphicx}
\usepackage{booktabs} 
\usepackage[bottom]{footmisc}

\def\ptFiguresDirectory#1{./figures/#1}
\def\FWER#1{FWER}
\def\fdr{\mathrm{FDR}}
\def\fnr{\mathrm{FNR}}
\def\fOne{F_{1}}
\def\neighCoeff{\gamma}
\def\dotProd#1#2{\left\langle {#1};{#2}\right\rangle}
\def\vnorm#1{\left\lVert {#1} \right\rVert}
\DeclareMathOperator{\sign}{sign}

\def\setBracks#1{\left\{{#1}\right\}}

\begin{document}
\begin{frontmatter}
\title{Combination of linear classifiers using score function -- analysis of possible combination strategies}
\author{Pawel Trajdos \and Robert Burduk}
\authorrunning{P. Trajdos \and R. Burduk}
\institute{Department of Systems and Computer Networks, Wroclaw
University of Science and Technology, \\ Wybrzeze Wyspianskiego 27, 50-370
Wroclaw, Poland \\ \texttt{pawel.trajdos@pwr.edu.pl}}

\maketitle

\begin{abstract}
In this work, we addressed the issue of combining linear classifiers using their score functions. The value of the scoring function depends on the distance from the decision boundary. Two score functions have been tested and four different combination strategies were investigated. During the experimental study, the proposed approach was applied to the heterogeneous ensemble and it was compared to two reference methods -- majority voting and model averaging respectively. The comparison was made in terms of seven different quality criteria. The result shows that combination strategies based on simple average, and trimmed average are the best combination strategies of the geometrical combination.
\end{abstract}
\end{frontmatter}

\section{Introduction}\label{sec:Introduction}

The combination of multiple base classifiers has been an important issue in machine learning for about twenty years~\cite{RB:drucker1994boosting},~\cite{RB:xu1992methods}. The ensembles of classifiers (EoC) or multiple classifiers systems (MCSs)~\cite{RB:Cyganek2012}, \cite{RB:markiewicz2015detection}, \cite{RB:Roli:PRL2001}, \cite{RB:przybyla2018three}, \cite{RB:wozniak2014survey} are popular in supervised classification algorithms where single classifiers are often unstable (small changes in input data may result in creation of very different decision boundaries) or are often more accurate than any of the base classifiers.

The task of constructing MCSs can be generally divided into three steps: generation, selection and integration~\cite{RB:Britto2014}. In the first step a set of base classifiers is trained using manipulation of the training patterns, manipulation of the training parameters or manipulation of    the feature space. 

The second phase of building EoCs is related to the choice of a set or one classifier from the whole available pool of base classifiers. It is popular to use the diversity measure to select one classifier or a subset of all base classifiers. In the literature, there are many approaches to the selection phase of building EoCs~\cite{RB:ko2008dynamic}, \cite{RB:burduk2015static}, \cite{rejer2017classifier}, \cite{RB:reif2014automatic}.

The integration process is the last stage of constructing EoCs and it is widely discussed in the pattern recognition literature~\cite{RB:ponti2011combining}, \cite{RB:tulyakov2008review}. Generally, supervised learning methods produce a classifier whose output is represented as a score function. This function is mapping to a function that is interpreted as a posteriori probability, rank level function or directly as a class label. Depending on the type of mapping, many methods for integrating base classifiers can be distinguished~\cite{kun2004}, \cite{RB:przybyla2017comparison}, \cite{RB:trawinski2017comparison}.

In this paper we propose the concept of the classifier integration process which uses score functions without their further transformation. In this paper we examined two forms of the score function that is called the potential function and four different combination strategies were investigated.

The remainder of this paper is organized as follows. Section~2 presents the proposed method of EoC integration using two types of the potential function. The experimental evaluation is presented in Section~3. The discussion and conclusions from the experiments are presented in Section~4.

\section{Proposed Method}\label{sec:ProposedMethod}
In this section, the proposed approach is explained. Additionally, this section introduces the notation used in this paper.
\subsection{Linear Binary Classifiers}\label{sec:ProposedMethod:BinLC}

In this paper, it is assumed that the input space $\mathbb{X}$ is a $d-\mathrm{dimensional}$ Euclidean space $\mathbb{X}=\mathbb{R}^d$. Each object from the input space $x\in\mathbb{X}$ belongs to one of two available classes, so the output space is: $\mathbb{M}=\left\{-1;1\right\}$. It is assumed that there exists an unknown mapping $f:\mathbb{X}\mapsto\mathbb{M}$ that assigns each input space coordinates into a proper class. A classifier $\psi:\mathbb{X}\mapsto\mathbb{M}$ is a function that is designed to provide an approximation of the unknown mapping $f$. A linear classifier makes its decision according to the following rule:
\begin{align}\label{eq:LinClass}
 \psi(x) &= \sign \left( \omega(x) \right),
\end{align}
where $\omega(x) = \dotProd{n}{x} + b$ is the so called \textit{discriminant function} of the classifier $\psi$~\cite{kun2004}, $n$ is a unit normal vector of the decision hyperplane ($\vnorm{n}=1$), $b$ is the distance from the hyperplane to the origin and $\dotProd{\cdot}{\cdot}$ is a dot product defined as follows:
\begin{align}
 \dotProd{a}{b} = \sum_{i=1}^{d}a_{i}b_{i},\;\forall a,b \in \mathbb{X}.
\end{align}
In this paper, we use a norm of the vector $x$ defined using the dot product:
\begin{align}\label{eq:dotP}
 \vnorm{x} &= \sqrt{\dotProd{x}{x}}.
\end{align}
When the normal vector of the plane is a unit vector, the absolute value of the discriminant function equals to the distance from the decision hyperplane to point $x$. The sign of the discriminant function depends on the site of the plane where the instance $x$ lies.

Now, let us define an ensemble classifier:
\begin{align}
  {\Psi} &=\setBracks{\psi^{(1)},\psi^{(2)},\cdots,\psi^{(N)}}
\end{align}
that is a set of $N$ classifiers that work together in order to produce a more robust result~\cite{kun2004}. In this paper, it is assumed that only linear, binary classifiers are employed. There are multiple strategies to combine the classifiers constituting the ensemble. The simplest strategy to combine the outcomes of multiple classifiers is to apply the majority voting scheme~\cite{kun2004}:
\begin{align}
 \label{eq:crispVoting}\omega_{\mathrm{MV}}(x) &= \sum_{i=1}^{N}\sign(\omega^{(i)}(x)),
\end{align}
where $\omega^{(i)}(x)$ is the value of the discriminant function provided by the classifier $\psi^{(i)}$ for point $x$. However, this simple yet effective strategy completely ignores the distance of the instance $x$ from the decision planes.

Another strategy is model averaging~\cite{Skurichina1998}. The output of the averaged model may be calculated by simply averaging the values of the discriminant functions:
\begin{align}
 \label{eq:softVoting} \omega_{\mathrm{MA}}(x) &= \frac{1}{N}\sum_{i=1}^{N}\omega^{(i)}(x)
\end{align}
After combining the base classifiers, the final prediction of the ensemble is obtained according to the rule~\eqref{eq:LinClass}.

\subsection{The Proposed Method}\label{sec:ProposedMethod:potential}

\begin{figure}[tb]
 \centering
  \includegraphics[width = 0.4\textwidth]{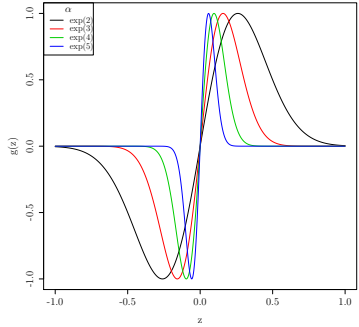}
  \caption{Potential function $g$.\label{fig:G2}}
\end{figure}

In this paper, an approach similar to the softmax~\cite{kun2004} normalization is proposed. Contrary to the softmax normalization, our goal is not to provide a probabilistic interpretation of the linear classifier but to provide a fusion technique that works in the geometrical space. The idea is to span a potential field around the decision plane. The potential field may be constructed by applying a transformation on the value of the discriminant function. The transformation must meet the following properties:
\begin{align}
 \label{eq:g:prop1} \sign(g(\omega^{(i)}(x))) &= \sign(\omega^{(i)}(x)) \forall x \in \mathbb{X},\\
 \label{eq:g:prop2} g(\omega^{(i)}(x)) &\in \left[-1;1 \right] \forall z \in \mathbb{R},\\
 \label{eq:g:prop3} g(0) &=0.
\end{align}
Property~\eqref{eq:g:prop1} assures that the crisp decision based on the transformed value is the same as the decision based on the unmodified discriminant function. Property~\eqref{eq:g:prop2} bounds $g$ in interval $[-1;1]$. However, contrary to the softmax normalization the transformation does not have to be a sigmoid function.
Property~\eqref{eq:g:prop3} assures that the potential is 0 at the surface of the decision plane.
In this paper, the following transformation function is used:
\begin{align}\label{eq:Potential2}
 g(z) &= z\exp(-\neighCoeff z^{2} +0.5)\sqrt{2\neighCoeff},
\end{align}
where $\neighCoeff$ is a coefficient that determines the position and steepness of the peak. The translation constant $0.5$ and the scaling factor $\sqrt{2\neighCoeff}$ guarantee that the maximum and minimum values are $1$ and $-1$
respectively. The function is visualised in the figure~\ref{fig:G2}.

All models in the ensemble share the same shape coefficient $\neighCoeff$. The shape coefficient is tuned in order to achieve the best quality of the entire ensemble.

After transforming the values of discriminant functions for the entire ensemble, there is a need to combine the outcomes to produce the final decision. In this paper, we analyze four different combination rules. The first one is a simple average of the transformed values of discriminant functions:
\begin{align}\label{eq:FunVotes}
 \omega_{\mathrm{TA}}(x) &= \frac{1}{N}\sum_{i=1}^{N}g(\omega^{(i)}(x)).
\end{align}

The other one is to apply the trimmed mean approach:
\begin{align}\label{eq:TMean}
 \omega_{\mathrm{TME}}(x) &= \frac{1}{N-2}\sum_{i=1}^{N} \left[g(\omega^{(i)}(x)) - \max_{\mathclap{i \in \{1,2,\cdots,N\}}}\omega^{(i)}(x) - \min_{\mathclap{i \in \{1,2,\cdots,N\}}}\omega^{(i)}(x)\right].
\end{align}

Before the remaining combination rules are defined, let us introduce subsets of negative and positive values of the transformed ensemble outcomes:
\begin{align}
 \mathcal{G}_{-}(x) &= \left\{ g(\omega^{(i)}(x)) \mid  g(\omega^{(i)}(x))<0  \right\},\\
 \mathcal{G}_{+}(x) &= \left\{ g(\omega^{(i)}(x)) \mid  g(\omega^{(i)}(x))\geq 0  \right\}.
\end{align}

Then, the remaining rules are as follows:
\begin{align}
 \label{eq:combMAX}\omega_{\mathrm{MAX}}(x) &= \max(\mathcal{G}_{+}(x)) +  \min(\mathcal{G}_{-}(x)),\\
 \label{eq:combMIN}\omega_{\mathrm{MIN}}(x) &= \min(\mathcal{G}_{+}(x)) +  \max(\mathcal{G}_{-}(x)),\\
 \label{eq:combMED}\omega_{\mathrm{GME}}(x) &= \left (\prod_{{z \in \mathcal{G}_{+}(x))}} \left\lvert z \right\lvert \right)^{{\left\lvert \mathcal{G}_{+}(x)) \right\lvert}^{-1}} - \left (\prod_{{z \in \mathcal{G}_{-}(x))}} \left\lvert z \right\lvert \right)^{{\left\lvert \mathcal{G}_{-}(x)) \right\lvert}^{-1}},
\end{align}
where $\left\lvert \mathcal{G}_{-}(x)) \right\lvert$ and $\left\lvert z \right\lvert$ are cardinality of set $\mathcal{G}_{-}(x))$ and the absolute value of $z$ respectively.

The proposed algorithm is able to deal only with the binary classification problems. However, any multi-class problem can be decomposed into multiple binary problems. In the experimental stage the One-vs-One strategy was used~\cite{Hllermeier2010}. This strategy builds a separate binary classifier for each pair of classes. In our method, a single pair-specific is replaced by the above-described ensemble classifier.

\section{Experimental Setup}\label{sec:ExpSetup}
In the conducted experimental study, the proposed approach was used to combine classifiers in the heterogeneous ensemble of classifier. The following base classifiers were employed:
\begin{itemize}
 \item $\psi_{\mathrm{FLDA}}$ -- Fisher LDA\cite{McLachlan1992}
 \item $\psi_{\mathrm{MLP}}$ -- single layer MLP classifier\cite{gurney1997an}
 \item $\psi_{\mathrm{NC}}$ -- nearest centroid (Nearest Prototype)\cite{manning2008,Kuncheva1998}
 \item $\psi_{\mathrm{SVM}}$ -- SVM classifier with linear kernel (no kernel)~\cite{Cortes1995},
 \item $\psi_{\mathrm{LR}}$ -- logistic regression classifier~\cite{Devroye1996}.
\end{itemize}
The classifiers implemented in WEKA framework~\cite{Hall2009} were used. The classifier parameters were set to their defaults. The multi-class problems were dealt with using One-vs-One decomposition~\cite{Hllermeier2010}. The experimental code was implemented using WEKA framework~\cite{Hall2009}.The source code of the algorithms is available online~\footnote{\url{https://github.com/ptrajdos/piecewiseLinearClassifiers/tree/master}}. The heterogeneous ensemble employs one copy of each of the above-mentioned base classifiers. Each classifier is learned using the entire dataset.

During the experimental evaluation the following combination methods were compared:
\begin{enumerate}
 \item $\Psi_{\mathrm{MV}}$ -- the ensemble combined using the majority voting approach,
 \item $\Psi_{\mathrm{MA}}$ -- the ensemble combined using the model averaging approach,
 \item $\Psi_{\mathrm{TA}}$ -- the ensemble combined using the rule described in~\eqref{eq:FunVotes}.
 \item $\Psi_{\mathrm{MAX}}$ -- the ensemble combined using the rule described in~\eqref{eq:combMAX}.
 \item $\Psi_{\mathrm{MIN}}$ -- the ensemble combined using the rule described in~\eqref{eq:combMIN}.
 \item $\Psi_{\mathrm{TME}}$ -- the ensemble combined using the rule described in~\eqref{eq:combMED}.
 \item $\Psi_{\mathrm{GME}}$ -- the ensemble combined using the rule described in~\eqref{eq:combMED}.
\end{enumerate}

The coefficient $\neighCoeff$ for transformation and $g$ was tuned using the grid search approach. The following set of parameter values were investigated:
\begin{align*}
 \left\{ \neighCoeff = \exp(i)| i \in\{2,\cdots,10\} \right\}.
\end{align*}
The parameter is chosen in such a way that it provides the maximum value of the macro-averaged $\fOne$ criterion.

To evaluate the proposed methods the following classification-quality criteria are used~\cite{Sokolova2009}: Zero-one loss (Accuracy); Macro-averaged $\fdr$, $\fnr$, $\fOne$;Micro-averaged $\fdr$, $\fnr$, $\fOne$.

Following the recommendations of~\cite{demsar2006} and~\cite{garcia2008extension}, the statistical significance of the obtained results was assessed using the two-step procedure. The first step is to perform the Friedman test~\cite{Friedman1940} for each quality criterion separately. Since the multiple criteria were employed, the familywise errors (\FWER{}) should be controlled~\cite{benjamini2001control}. To do so, the Bergman-Hommel~\cite{Bergmann1988} procedure of controlling \FWER{} of the conducted Friedman tests was employed. When the Friedman test shows that there is a significant difference within the group of classifiers, the pairwise tests, which use the Wilcoxon signed-rank test~\cite{wilcoxon1945}, \cite{demsar2006} were employed. To control \FWER{} of the Wilcoxon-testing procedure, the Bergman-Hommel approach was employed~\cite{holm1979}. For all the tests the significance level was set to $\alpha=0.05$.

Table~\ref{tab:BenchmarkSetsCharacteristics} displays the collection of the $64$ benchmark sets that were used during the experimental evaluation of the proposed algorithms. The table is divided into two columns.  Each column is organized as follows. The first column contains the names of the datasets. The remaining ones contain the set-specific characteristics of the benchmark sets: the number of instances in the dataset ($|S|$); dimensionality of the input space ($d$); the number of classes ($C$);average imbalance ratio ($\mathrm{IR}$).

The datasets come from the Keel~\footnote{\url{https://sci2s.ugr.es/keel/category.php?cat=clas}} repository or are generated by us.
The datasets are available online~\footnote{\url{https://github.com/ptrajdos/MLResults/blob/master/data/slDataFull.zip}}.

During the dataset-preprocessing stage, a few transformations on datasets were applied. That is, features are selected using the correlation-based approach~\cite{Hall1999}. Then, the PCA method was applied~\cite{Pearson1901} and the percentage of variance was set to $0.95$. The attributes were also scaled to fit the interval $[0;1]$. Additionally, in order to ensure the dot product to be in the interval $\left[-1;1\right]$, vectors in each dataset were scaled using the factor $\frac{1}{d^2}$. This normalization makes it easier to find proper $\neighCoeff$.

{
\setlength\tabcolsep{2.0pt}%
\def\arraystretch{0.9}%
\begin{table*}
 \centering\tiny
 \caption{The characteristics of the benchmark sets}\label{tab:BenchmarkSetsCharacteristics}
 \begin{tabular}{lcccc|lcccc|lcccc}
Name&$|S|$&$d$&$C$&$\mathrm{IR}$&Name&$|S|$&$d$&$C$&$\mathrm{IR}$&Name&$|S|$&$d$&$C$&$\mathrm{IR}$\\
\hline
appendicitis&106&7&2&2.52&housevotes&435&16&2&1.29&shuttle&57999&9&7&1326.03\\
australian&690&14&2&1.12&ionosphere&351&34&2&1.39&sonar&208&60&2&1.07\\
balance&625&4&3&2.63&iris&150&4&3&1.00&spambase&4597&57&2&1.27\\
banana2D&2000&2&2&1.00&led7digit&500&7&10&1.16&spectfheart&267&44&2&2.43\\
bands&539&19&2&1.19&lin1&1000&2&2&1.01&spirals1&2000&2&2&1.00\\
Breast Tissue&105&9&6&1.29&lin2&1000&2&2&1.83&spirals2&2000&2&2&1.00\\
check2D&800&2&2&1.00&lin3&1000&2&2&2.26&spirals3&2000&2&2&1.00\\
cleveland&303&13&5&5.17&magic&19020&10&2&1.42&texture&5500&40&11&1.00\\
coil2000&9822&85&2&8.38&mfdig fac&2000&216&10&1.00&thyroid&7200&21&3&19.76\\
dermatology&366&34&6&2.41&movement libras&360&90&15&1.00&titanic&2201&3&2&1.55\\
diabetes&768&8&2&1.43&newthyroid&215&5&3&3.43&twonorm&7400&20&2&1.00\\
Faults&1940&27&7&4.83&optdigits&5620&62&10&1.02&ULC&675&146&9&2.17\\
gauss2DV&800&2&2&1.00&page-blocks&5472&10&5&58.12&vehicle&846&18&4&1.03\\
gauss2D&4000&2&2&1.00&penbased&10992&16&10&1.04&Vertebral Column&310&6&3&1.67\\
gaussSand2&600&2&2&1.50&phoneme&5404&5&2&1.70&wdbc&569&30&2&1.34\\
gaussSand&600&2&2&1.50&pima&767&8&2&1.44&wine&178&13&3&1.23\\
glass&214&9&6&3.91&ring2D&4000&2&2&1.00&winequality-red&1599&11&6&20.71\\
haberman&306&3&2&1.89&ring&7400&20&2&1.01&winequality-white&4898&11&7&82.94\\
halfRings1&400&2&2&1.00&saheart&462&9&2&1.44&wisconsin&699&9&2&1.45\\
halfRings2&600&2&2&1.50&satimage&6435&36&6&1.66&yeast&1484&8&10&17.08\\
hepatitis&155&19&2&2.42&Seeds&210&7&3&1.00&&&&&\\
HillVall&1212&100&2&1.01&segment&2310&19&7&1.00&&&&&\\
\end{tabular}
\end{table*}
}

\section{Results and Discussion}\label{sec:Results}

\begin{figure}[tb]
 \centering
  \includegraphics[width = 0.7\textwidth]{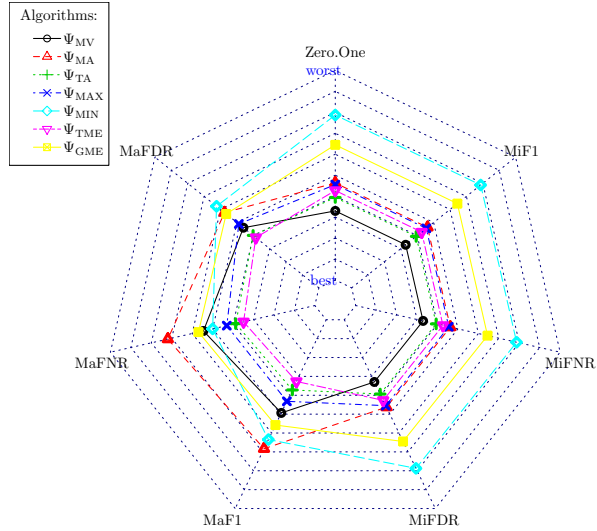}
  \caption{Average ranks of for the heterogeneous ensemble.\label{fig:Hetero}}
\end{figure}

To compare multiple algorithms on multiple benchmark sets the average ranks approach~\cite{demsar2006} is used. In the approach, the winning algorithm achieves rank equal '1', the second achieves rank equal '2', and so on. In the case of ties, the ranks of algorithms that achieve the same results, are averaged. To provide a visualisation of the average ranks, the radar plots are employed. In the plots, the data is visualised in such a way that the lowest ranks are closer to the centre of the graph. The radar plots related to the experimental results are shown in figure~\ref{fig:Hetero}.

Due to the page limit, the full results are published online~\footnote{\url{https://github.com/ptrajdos/MLResults/blob/master/Boundaries/bounds_hetero_15.01.2019E4_m_R.zip}}

The numerical results are given in Table \ref{table:Hetero}. The table is structured as follows. The first row contains names of the investigated algorithms. Then, the table is divided into seven sections -- one section is related to a single evaluation criterion. The first row of each section is the name of the quality criterion investigated in the section. The second row shows the p-value of the Friedman test. The third one shows the average ranks achieved by algorithms. The following rows show p-values resulting from pairwise Wilcoxon test. The p-value which is equal to $0.000$ informs that the p-values are lower than $10^{-3}$ and p-value is equal to $1.000$ informs that the value is higher than $0.999$.

The analysis of the radar plot suggests that two groups of classification criteria can be distinguished. The first group contains micro-averaged criteria and the zero-one criterion, the second one contains macro-averaged criteria. Evaluation of the classifiers carried out with the use of criteria belonging to a specific group reveals different relationships between classifiers. These differences are a consequence of the properties of the quality criteria used. This means that the zero-one criterion and micro-averaged criteria give us information related to the classification quality for the majority classes. On the other hand, the macro-averaged criteria put more emphasis on classification quality for minority classes~\cite{Sokolova2009}.

For the zero-one criterion and micro-averaged criteria, three main groups of classifiers can be seen. The first group contains $\Psi_{\mathrm{MIN}}$ and $\Psi_{\mathrm{GME}}$ classifiers that perform significantly worse than the other analysed classifiers. What is more, classifier $\Psi_{\mathrm{MIN}}$ is significantly worse than $\Psi_{\mathrm{GME}}$ for all quality criteria belonging to the investigated group. The second group contains only one classifier -- $\Psi_{\mathrm{MV}}$. According to average ranks, this classifier is the best performing one for the investigated set of quality criteria. According to the statistical analysis, this classifier outperforms the remaining classifiers except for $\Psi_{\mathrm{TA}}$ and $\Psi_{\mathrm{TME}}$. The third group consisted of classifiers $\Psi_{\mathrm{MA}}$, $\Psi_{\mathrm{TA}}$, $\Psi_{\mathrm{MAX}}$, and $\Psi_{\mathrm{TME}}$. There are no significant differences between the classifiers within this group.

For macro-averaged measures, the situation changes significantly. First of all, it may be noticed that average ranks of reference methods ($\Psi_{\mathrm{MV}}$ and $\Psi_{\mathrm{MA}}$) increase, whereas the average ranks of the proposed methods decrease. That is, the model-averaging classifier $\Psi_{\mathrm{MA}}$ becomes the worst one except for $\Psi_{\mathrm{MIN}}$ according to macro-averaged $F_1$ and FNR criteria. The majority voting classifier $\Psi_{\mathrm{MV}}$ also deteriorates significantly. Now it is comparable to $\Psi_{\mathrm{MAX}}$, $\Psi_{\mathrm{MIN}}$ and $\Psi_{\mathrm{GME}}$ classifiers. What is more, $\Psi_{\mathrm{MV}}$ classifier is outperformed by $\Psi_{\mathrm{TA}}$ and $\Psi_{\mathrm{TME}}$ classifiers in terms of macro-averaged FNR and $F_1$ criteria. The reason for the above-mentioned deterioration of the reference methods is the fact that they are not tuned to perform better on minority classes, whereas the investigated methods were tuned to do so.

Now let us investigate the differences inside the group of the proposed combination criteria. First of all, classifiers $\Psi_{\mathrm{TA}}$ and $\Psi_{\mathrm{TME}}$ offer the best classification quality under macro-averaged $F_1$ measure. It means that these classifiers offer the best trade-off between macro-averaged precision and recall. Under macro-averaged FDR ($1-\mathrm{precision}$) measure, these algorithms outperform only $\Psi_{\mathrm{MIN}}$ and $\Psi_{\mathrm{GME}}$ classifiers. For macro-averaged FNR ($1-\mathrm{recall}$) the investigated classifiers outperform all but $\Psi_{\mathrm{MIN}}$ classifiers. On the other hand, under the macro-averaged measures, there are no significant differences between $\Psi_{\mathrm{TA}}$ and $\Psi_{\mathrm{TME}}$.

{
\setlength\tabcolsep{1.0pt}%
\begin{table}[ht]
\centering\tiny
\caption{Statistical evaluation. Wilcoxon test for the heterogeneous ensemble -- p-values for paired comparisons of the investigated methods.\label{table:Hetero}}
\begin{tabular}{c|ccccccc|ccccccc|ccccccc}
  & \scalebox{0.75}{$\Psi_{\mathrm{MV}}$} & \scalebox{0.75}{$\Psi_{\mathrm{MA}}$} & \scalebox{0.75}{$\Psi_{\mathrm{TA}}$} & \scalebox{0.75}{$\Psi_{\mathrm{MAX}}$} & \scalebox{0.75}{$\Psi_{\mathrm{MIN}}$} & \scalebox{0.75}{$\Psi_{\mathrm{TME}}$} & \scalebox{0.75}{$\Psi_{\mathrm{GME}}$} & \scalebox{0.75}{$\Psi_{\mathrm{MV}}$} & \scalebox{0.75}{$\Psi_{\mathrm{MA}}$} & \scalebox{0.75}{$\Psi_{\mathrm{TA}}$} & \scalebox{0.75}{$\Psi_{\mathrm{MAX}}$} & \scalebox{0.75}{$\Psi_{\mathrm{MIN}}$} & \scalebox{0.75}{$\Psi_{\mathrm{TME}}$} & \scalebox{0.75}{$\Psi_{\mathrm{GME}}$} & \scalebox{0.75}{$\Psi_{\mathrm{MV}}$} & \scalebox{0.75}{$\Psi_{\mathrm{MA}}$} & \scalebox{0.75}{$\Psi_{\mathrm{TA}}$} & \scalebox{0.75}{$\Psi_{\mathrm{MAX}}$} & \scalebox{0.75}{$\Psi_{\mathrm{MIN}}$} & \scalebox{0.75}{$\Psi_{\mathrm{TME}}$} & \scalebox{0.75}{$\Psi_{\mathrm{GME}}$} \\
  \hline
Nam&\multicolumn{7}{c|}{Zero-One}&\multicolumn{7}{c|}{MaFDR}&\multicolumn{7}{c}{MaFNR}\\
Frd&\multicolumn{7}{c|}{5.729e-14}&\multicolumn{7}{c|}{2.873e-04}&\multicolumn{7}{c}{1.791e-08}\\
Rnk & 2.98 & 3.78 & 3.36 & 3.73 & 5.72 & 3.56 & 4.87 & 3.76 & 4.45 & 3.41 & 3.93 & 4.75 & 3.31 & 4.39 & 4.27 & 5.32 & 3.32 & 3.58 & 4.00 & 3.09 & 4.42 \\
   \cmidrule(lr){2-8}\cmidrule(lr){9-15}\cmidrule(lr){16-22}
  \scalebox{0.75}{$\Psi_{\mathrm{MV}}$} &  & .007 & .091 & .002 & .000 & .161 & .000 &  & .016 & .295 & .969 & .155 & .279 & .673 &  & .000 & .003 & .505 & 1.00 & .000 & 1.00 \\
  \scalebox{0.75}{$\Psi_{\mathrm{MA}}$} &  &  & .968 & .968 & .000 & .968 & .007 &  &  & .001 & .025 & .878 & .002 & .295 &  &  & .000 & .000 & .018 & .000 & .002 \\
  \scalebox{0.75}{$\Psi_{\mathrm{TA}}$} &  &  &  & .080 & .000 & .968 & .000 &  &  &  & .056 & .013 & .878 & .025 &  &  &  & .049 & .139 & 1.00 & .008 \\
  \scalebox{0.75}{$\Psi_{\mathrm{MAX}}$} &  &  &  &  & .000 & .846 & .000 &  &  &  &  & .028 & .056 & .155 &  &  &  &  & .601 & .049 & .016 \\
  \scalebox{0.75}{$\Psi_{\mathrm{MIN}}$} &  &  &  &  &  & .000 & .000 &  &  &  &  &  & .004 & .295 &  &  &  &  &  & .139 & 1.00 \\
  \scalebox{0.75}{$\Psi_{\mathrm{TME}}$} &  &  &  &  &  &  & .000 &  &  &  &  &  &  & .003 &  &  &  &  &  &  & .001 \\
   \hline
Nam&\multicolumn{7}{c|}{MaF1}&\multicolumn{7}{c|}{MiFDR}&\multicolumn{7}{c}{MiFNR}\\
Frd&\multicolumn{7}{c|}{2.641e-09}&\multicolumn{7}{c|}{5.729e-14}&\multicolumn{7}{c}{5.729e-14}\\
Rnk & 3.96 & 5.10 & 3.23 & 3.59 & 4.81 & 2.96 & 4.35 & 2.98 & 3.78 & 3.36 & 3.73 & 5.72 & 3.56 & 4.87 & 2.98 & 3.78 & 3.36 & 3.73 & 5.72 & 3.56 & 4.87 \\
    \cmidrule(lr){2-8}\cmidrule(lr){9-15}\cmidrule(lr){16-22}
  \scalebox{0.75}{$\Psi_{\mathrm{MV}}$} &  & .000 & .017 & .548 & .117 & .000 & .340 &  & .007 & .091 & .002 & .000 & .161 & .000 &  & .007 & .091 & .002 & .000 & .161 & .000 \\
  \scalebox{0.75}{$\Psi_{\mathrm{MA}}$} &  &  & .000 & .000 & .315 & .000 & .017 &  &  & .968 & .968 & .000 & .968 & .007 &  &  & .968 & .968 & .000 & .968 & .007 \\
  \scalebox{0.75}{$\Psi_{\mathrm{TA}}$} &  &  &  & .017 & .002 & .454 & .001 &  &  &  & .080 & .000 & .968 & .000 &  &  &  & .080 & .000 & .968 & .000 \\
  \scalebox{0.75}{$\Psi_{\mathrm{MAX}}$} &  &  &  &  & .007 & .014 & .011 &  &  &  &  & .000 & .846 & .000 &  &  &  &  & .000 & .846 & .000 \\
  \scalebox{0.75}{$\Psi_{\mathrm{MIN}}$} &  &  &  &  &  & .000 & .185 &  &  &  &  &  & .000 & .000 &  &  &  &  &  & .000 & .000 \\
  \scalebox{0.75}{$\Psi_{\mathrm{TME}}$} &  &  &  &  &  &  & .000 &  &  &  &  &  &  & .000 &  &  &  &  &  &  & .000 \\
   \hline
Nam&\multicolumn{7}{c|}{MiF1}&\multicolumn{7}{c|}{}&\multicolumn{7}{c}{}\\
Frd&\multicolumn{7}{c|}{5.729e-14}&\multicolumn{7}{c|}{}&\multicolumn{7}{c}{}\\
Rnk & 2.98 & 3.78 & 3.36 & 3.73 & 5.72 & 3.56 & 4.87 &  &  &  &  &  &  &  &  &  &  &  &  &  &  \\
    \cmidrule(lr){2-8}
  \scalebox{0.75}{$\Psi_{\mathrm{MV}}$} &  & .007 & .091 & .002 & .000 & .161 & .000 &  &  &  &  &  &  &  &  &  &  &  &  &  &  \\
  \scalebox{0.75}{$\Psi_{\mathrm{MA}}$} &  &  & .968 & .968 & .000 & .968 & .007 &  &  &  &  &  &  &  &  &  &  &  &  &  &  \\
  \scalebox{0.75}{$\Psi_{\mathrm{TA}}$} &  &  &  & .080 & .000 & .968 & .000 &  &  &  &  &  &  &  &  &  &  &  &  &  &  \\
  \scalebox{0.75}{$\Psi_{\mathrm{MAX}}$} &  &  &  &  & .000 & .846 & .000 &  &  &  &  &  &  &  &  &  &  &  &  &  &  \\
  \scalebox{0.75}{$\Psi_{\mathrm{MIN}}$} &  &  &  &  &  & .000 & .000 &  &  &  &  &  &  &  &  &  &  &  &  &  &  \\
  \scalebox{0.75}{$\Psi_{\mathrm{TME}}$} &  &  &  &  &  &  & .000 &  &  &  &  &  &  &  &  &  &  &  &  &  &  \\
  \end{tabular}
\end{table}
}

\section{Conclusions}\label{sec:Conclusions}

In this paper, a geometric combination scheme was proposed. Four different methods of producing the final output of EoC were investigated. The goal of this paper is to determine the best combination strategy for the given potential-function-induced geometrical space. The experimental comparison shows that $\Psi_{\mathrm{TA}}$ and $\Psi_{\mathrm{TME}}$ algorithms are the best choice. This is because under macro-averaged measures they are outperforming the other proposed strategies and reference methods. What is more, under the micro-averaged criteria they are comparable to the majority voting procedure. According to the outcome of the statistical evaluation, these algorithms perform equally well. However, under macro-averaged measures, $\Psi_{\mathrm{TME}}$ achieves a slightly lower average rank. This suggests that $\Psi_{\mathrm{TME}}$ may be slightly better since the truncated mean combination rule removes extreme values of the potential function so it may be less influenced by outliers.

The obtained results are very interesting, so we are willing to continue our research in the field of combining classifiers in the geometrical space. An interesting direction to explore may be the application of the potential function whose shape is not given arbitrary but is created considering data distribution.

\textbf{Acknowledgments.} This work was supported in part by the National Science Centre, Poland under the grant no. 2017/25/B/ST6/01750.

 \bibliography{bibliography}

\end{document}